\documentclass[sigconf]{acmart}

\usepackage{makecell}
\usepackage[flushleft]{threeparttable}
\usepackage{tablefootnote}

\AtBeginDocument{%
  \providecommand\BibTeX{{%
    \normalfont B\kern-0.5em{\scshape i\kern-0.25em b}\kern-0.8em\TeX}}}

\begin{document}

\title[RecipeGPT: Cooking Recipe Generation and Evaluation System]{RecipeGPT: Generative Pre-training Based Cooking Recipe Generation and Evaluation System}

\author{Helena H. Lee}
\affiliation{%
  \institution{Singapore Management University}
}
\email{helenalee@smu.edu.sg}

\author{Ke Shu}
\affiliation{%
  \institution{Singapore Management University}
}
\email{keshu@smu.edu.sg}

\author{Palakorn Achananuparp}
\affiliation{%
  \institution{Singapore Management University}
}
\email{palakorna@smu.edu.sg}

\author{Philips Kokoh Prasetyo}
\affiliation{%
  \institution{Singapore Management University}
}
\email{pprasetyo@smu.edu.sg}

\author{Yue Liu}
\affiliation{%
  \institution{Singapore Management University}
}
\email{yueliu@smu.edu.sg}

\author{Ee-Peng Lim}
\affiliation{%
  \institution{Singapore Management University}
}
\email{eplim@smu.edu.sg}

\author{Lav R. Varshney}
\affiliation{%
  \institution{University of Illinois at Urbana-Champaign}
}
\email{varshney@illinois.edu}

\renewcommand{\shortauthors}{Lee et al.}

\begin{abstract}

Interests in the automatic generation of cooking recipes have been growing steadily over the past few years thanks to a large amount of online cooking recipes. We present RecipeGPT, a novel online recipe generation and evaluation system. The system provides two modes of text generations: (1) instruction generation from given recipe title and ingredients; and (2) ingredient generation from recipe title and cooking instructions. Its back-end text generation module comprises a generative pre-trained language model GPT-2 fine-tuned on a large cooking recipe dataset. Moreover, the recipe evaluation module allows the users to conveniently inspect the quality of the generated recipe contents and store the results for future reference. 
RecipeGPT can be accessed online at ~\url{https://recipegpt.org/}




\end{abstract}


\setlength{\textfloatsep}{1.5pt plus 1pt minus 1pt}
\setlength{\abovecaptionskip}{1pt plus 1pt minus 0pt}
\setlength{\belowcaptionskip}{1pt plus 1pt minus 0pt}

\maketitle
\section{Introduction}

Automatic generation of cooking recipes is an interesting and practical research problem that can help overcome the limitations of standard recipe retrieval systems. Though many online recipe databases, such as Allrecipes and Yummly, allow users to explicitly include and exclude specific ingredients in the recipe search, the users have to use an advanced search interface which can be difficult to understand for novice users. Recipe generation systems can facilitate this process by directly generating cooking instructions for a specific recipe given a list of user-specified ingredients. Next, it can also be used for creative cooking (e.g., IBM Chef Watson \cite{varshney2019big}), where the system helps the users to come up with novel and practical ways for cooking certain dishes, taking into account the complementarity of ingredients.

\begin{figure*}[tp] 
\centering
\scalebox{0.85}{
\includegraphics[width=\linewidth]{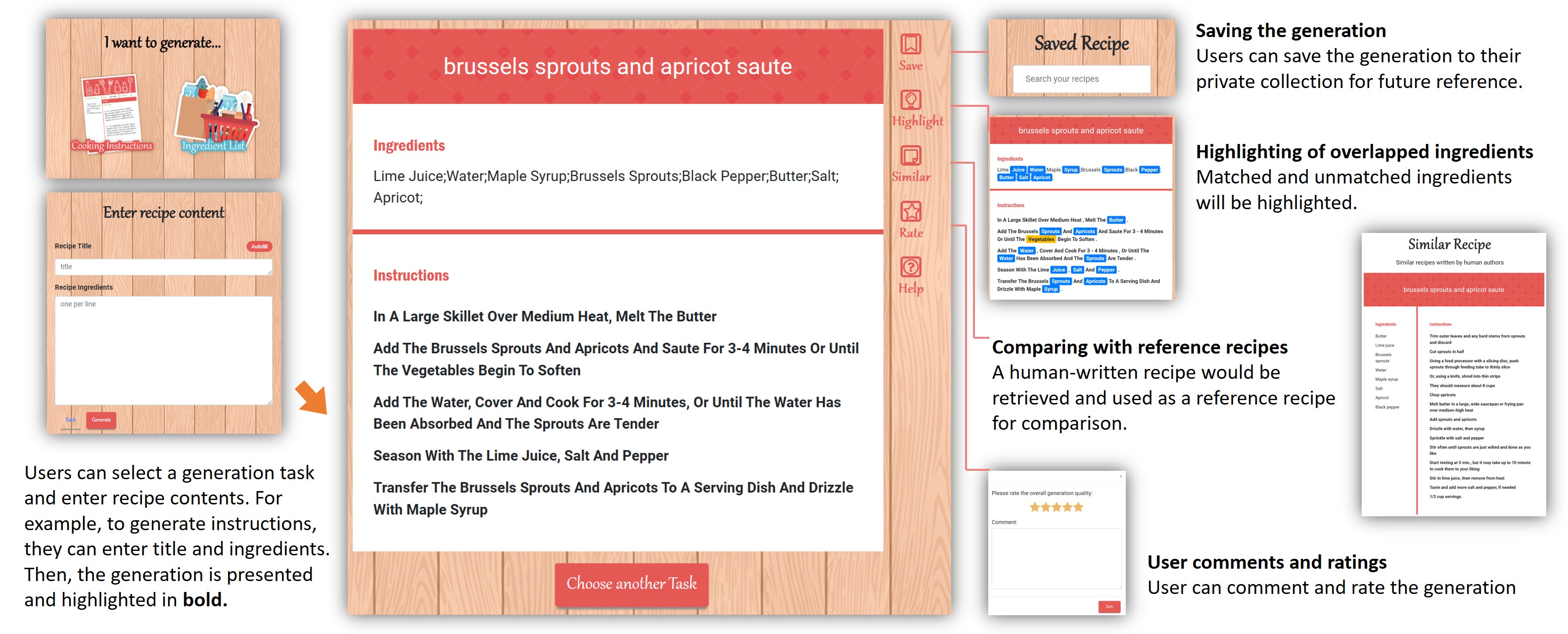}
}
\caption{Overview of RecipeGPT}
\label{fig:userflow}
\end{figure*}

A few approaches to recipe text generation have been proposed, such as knowledge-based models \cite{varshney2019big} and deep neural networks models \cite{bosselut2017simulating,majumder2019generating,salvador2019inverse}. Large-scale transformer-based language models have been shown to outperform Recurrent Neural Networks (RNNs) in several natural language processing (NLP) tasks lately. In text generation, transformers are known for their effectiveness in capturing complex dependencies and generating fluent sentences. Among those, OpenAI’s GPT-2, pre-trained on a gigaword-scale textual dataset, has achieved impressive results in a variety of text generation tasks~\cite{radford2019language}. Recent study has also shown that fine-tuning GPT-2 can result in better performance on domain-specific text generations~\cite{zhang2019dialogpt}. 
However, the effectiveness of pre-trained transformer-based language models in cooking recipe generation has not yet been explored. 


Similar to other text generation tasks, evaluating the quality of machine-generated recipe texts is challenging. First, most neural text generation models are non-deterministic, thus each generation run produces unique results which are difficult to replicate. Second, machine-evaluation metrics and human evaluation on text generations are not well correlated. Third, a lot of efforts are required to judge the content coverage in the generated text. Finally, the adaptability of the model has not been well-studied, for example, generating cooking instructions from novel inputs of recipe title and ingredient combinations.


\begin{figure}[htp] 
\centering
\scalebox{1}{
\includegraphics[width=0.9\linewidth]{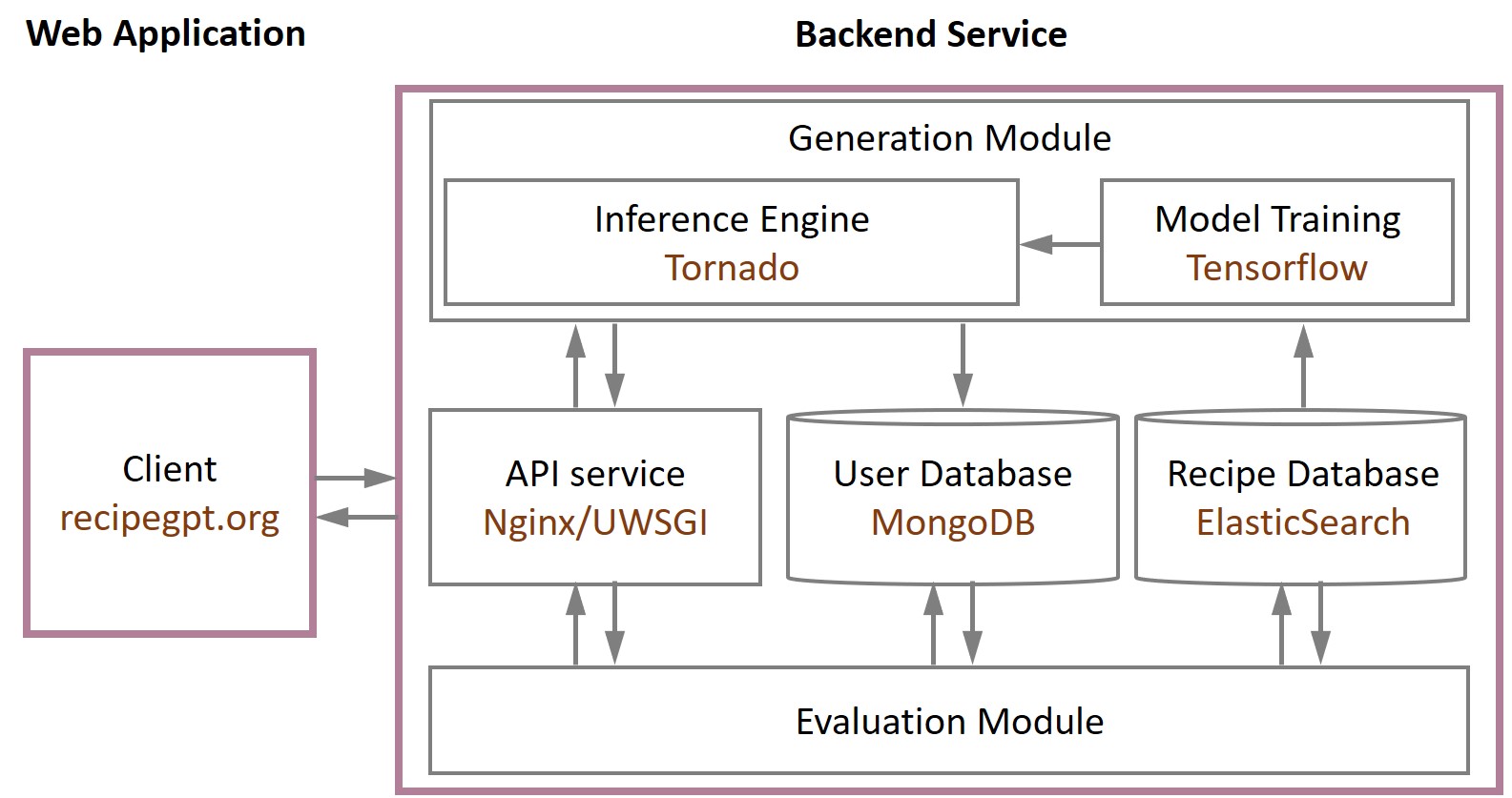}
}
\caption{System Architecture}
\label{fig:arch}
\end{figure}

In this paper, we introduce RecipeGPT, a novel web application for recipe generation and evaluation, to demonstrate the feasibility of generative pre-trained transformer in cooking recipe generation and to assist users in evaluating the generation quality more easily as illustrated in Figure~\ref{fig:userflow}. 
Users can utilize RecipeGPT to: 1) generate cooking instructions according to given recipe title and ingredient texts; and 2) generate ingredients given recipe title and cooking instruction texts. The system allows users to rate, comment, and store the generated results. Furthermore, the system also helps the users compare machine-generated recipes with similar human-written recipes retrieved from the recipe database.

\section{System Overview}

Figure \ref{fig:arch} shows the system architecture of RecipeGPT. First, the user client is developed as a web application (presented in Figure \ref{fig:userflow}). Next, the backend service provides core RESTFul web services which can be used by clients with an API key. These services include accepting user inputs, handling user interactions, and returning the generated results. More specifically, the backend service is composed of two following modules:


\subsection{Generation Module}
The generation module relies on a generative pre-trained transformer GPT-2, fine-tuned on Recipe1M dataset \cite{marin2019recipe1m}. We utilize the fine-tuned model to perform two tasks: ingredient generation and instruction generation. 
The details of model training are described in Section~\ref{sec:model}. 
The generation model is trained using Tensorflow and deployed to the inference engine to handle live requests. Then, utilizing an API service built upon Nginx and UWSGI server, the requests are encapsulated and sent to the inference engine accelerated by GPU to increase the request throughput. Lastly, MongoDB is used as a data storage to optionally save the generated outputs for future reference and inspection.

\subsection{Evaluation Module}
The aim of the evaluation module is to provide functionalities to assist the users in individually inspecting the quality of the generated recipe texts. These are implemented in the following features:

\textbf{Highlighting of overlapped ingredients.} To quickly check the quality of the generated recipe texts, users may want to compare the overlap of ingredient words in the specified recipe contexts and the generated texts. High-quality generations are those that have the highest degree of overlapped ingredients, i.e., all ingredient words specified by the users appear in the generated texts. RecipeGPT facilitates the inspection of overlapped ingredients through a word highlighting feature. Since recipe authors tend to use different word variations to refer to the same ingredients (e.g., simply \textit{cheese} instead of \textit{provolone cheese}), we only consider root nouns of ingredients in the comparison.


\textbf{Comparing with reference recipes.} For each generation run, RecipeGPT also retrieves the most similar recipe given the specified recipe contexts (e.g., title, ingredients, etc.) from Recipe1M, stored in ElasticSearch data storage, to be used as a reference to compare against the generated recipe texts. We use ElasticSearch's built-in search functions for ranking recipes by similarity.


\textbf{User comments and ratings.} Lastly, ReciptGPT provides basic annotation functions, such as user commenting and rating, to facilitate human evaluation of the generated recipe texts. The annotation data are stored together with the user-saved recipes.

\section{Training Recipe Generation Model} \label{sec:model}

In this section, we describe in detail the steps to build our recipe generative pre-training transformer, one of the core components of the recipe generation module, by fine-tuning GPT-2 on Recipe1M dataset. These involve (1) preprocessing of Recipe1M data; and (2) training multi-field recipe generation model.


\subsection{Data Preprocessing} 
First, we filter out numerals, quantity words, and other comment texts in the ingredients using an ingredient phrase parser\footnote{\url{https://github.com/nytimes/ingredient-phrase-tagger}} and our own rule-based filtering. In addition, we remove recipes that contain non-ingredients and non-instructional sentences (e.g., nutritional or author information). Finally, we keep the remaining recipes with at least 2 ingredients, 2 instructional sentences, and 20 words in the instructions (N = 904,401). From those, two disjointed sets of 4,000 recipes each are reserved for the validation set and the testing set.

\begin{figure}[ht] 
\centering
\scalebox{0.9}{
\includegraphics[width=\linewidth]{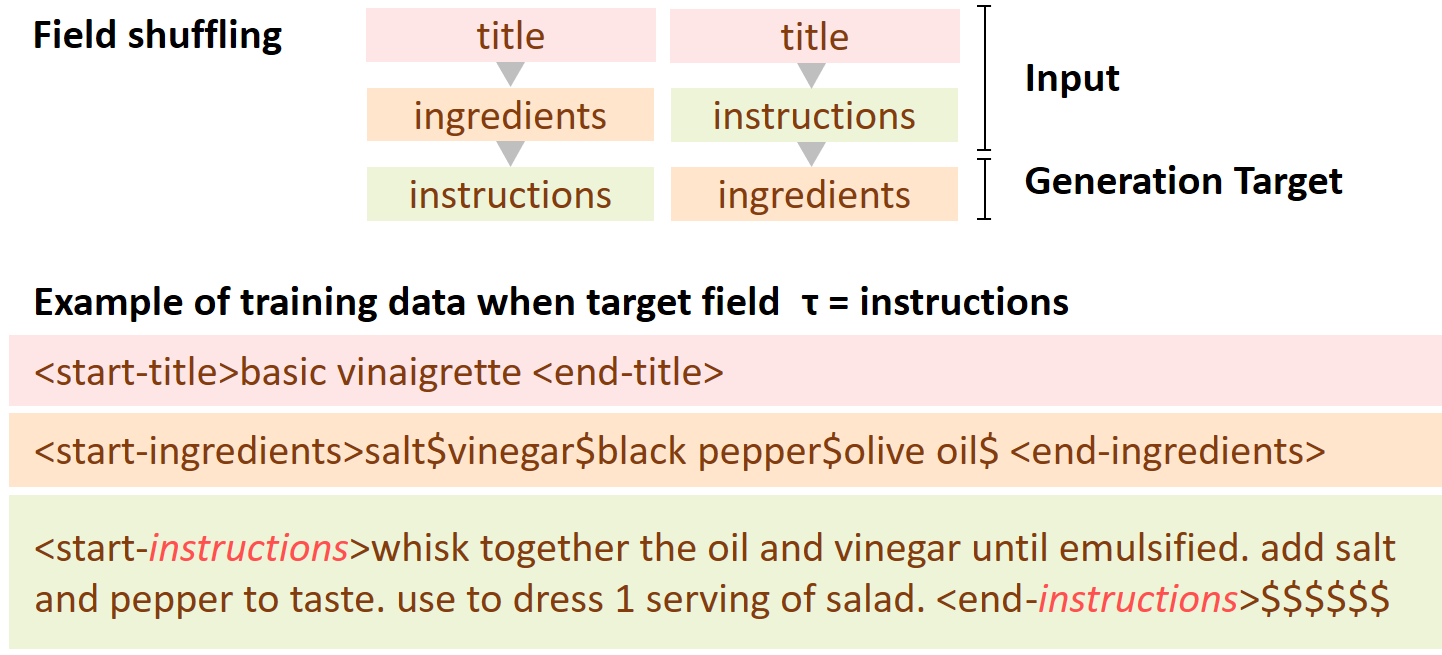}
}
\caption{Illustration of Multi-field Generation}
\label{fig:shuffling}
\end{figure}

\subsection{Multi-field Learning and Generation}
Our aim is to train GPT-2 transformer provided by Radford et al \cite{radford2019language} to generate multi-field recipe documents. Specifically, each recipe in Recipe1M consists of three fields: title, ingredients, and cooking instructions.
To that end, we follow Zeller et al.'s approach~\cite{zellers2019defending} to perform multi-field learning and generation. Each field is encapsulated by field-specific start and end tokens. From the original training set, we further constructed a multi-field training set by shuffling the input and the target fields. We also shuffle the order of ingredients. Figure~\ref{fig:shuffling} illustrates how we shuffle the training data to enable the generation of target fields given recipe contexts as input fields. 
 
 To generate a target field $\tau$, we append the field-specific start token <start-$\tau$> to the input followed by sampling from the model until we hit <end-$\tau$>. Note that we might have padding tokens `\$' after <end-$\tau$>. During each sampling step, we select a token among the $k$ highest-ranked tokens of the entire vocabulary sorted by the prediction probability. This sampling strategy is also known as top-$k$ sampling. Although the generation process is non-deterministic, we can control the generation diversity by the hyperparameter $k$. The higher the $k$, the larger the diversity of generated texts.

We select the optimal learning rate based on perplexity using the validation set. Next, we follow the Byte-Pair Encoding used in GPT-2 to process the inputs. The maximum number of tokens is set to 512 since it fully covers 98.6\% recipes in Recipes1M. We add padding tokens at the end of each recipe to unify the sampling length. For recipes exceeding maximum tokens, a randomly selected chunk of 512 tokens is sampled in each training iteration. Due to the memory limitation in our training environment (16GB), we set the batch size to 8. 


\section{Experiments} \label{sec:results}
In this section, we describe the experimental setup to evaluate the performance of the fine-tuned recipe GPT-2 models on the 4,000 recipe data in the test set. Specifically, two recipe text generation tasks are focused: (1) generating ingredient texts from recipe title and cooking instructions; and (2) generating cooking instruction texts from title and ingredients. We try to find the best sampling hyperparameters and select the most suitable model for live deployment. Lastly, we further examine the degree of coherence of ingredients mentioned in different fields within the same recipes. A high-quality recipe should consistently refer to the same set of ingredients across all fields.



\begin{figure}[htp] 
\centering
\scalebox{0.9}{
\includegraphics[width=\linewidth]{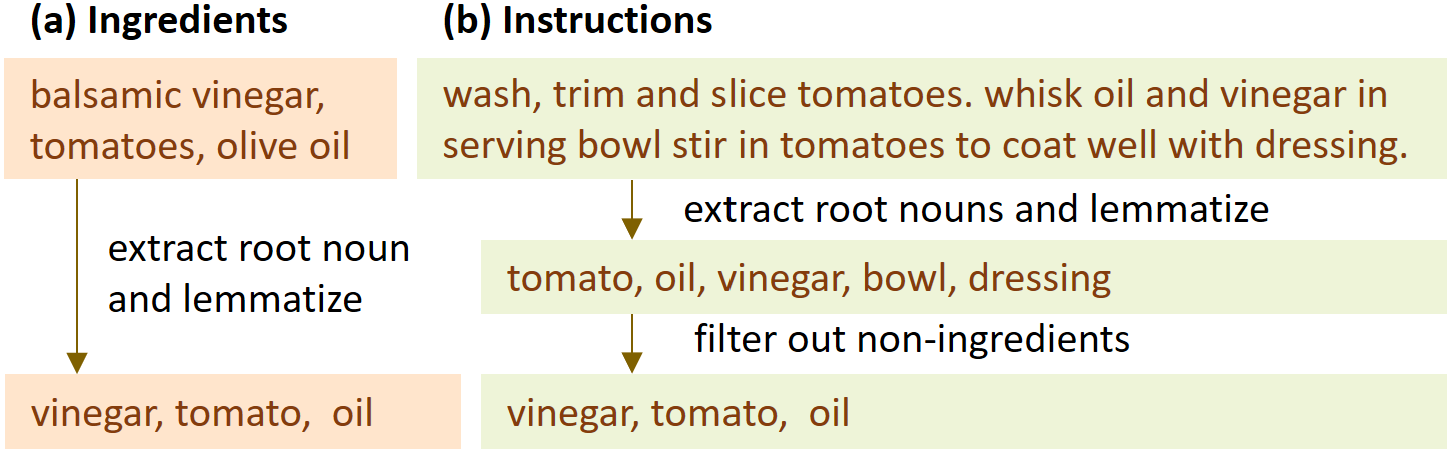}
}
\caption{Extracting Ingredient-containing Root Nouns}
\label{fig:spacy}
\end{figure}

\subsection{Evaluation Metrics}  \label{sec:metrics}
\textbf{Ingredient generation.} 
We compute standard F1 scores between the set of generated ingredients and ground-truth ingredients. More specifically, we only consider lemmatized root nouns of ingredients as units of analysis instead of the whole noun phrases in the evaluation. we utilize spaCy\footnote{\url{https://spacy.io}} to identify the lemmatized root nouns as illustrated in Figure \ref{fig:spacy}(a).

\textbf{Instruction generation.} 
We compute two standard n-gram overlap based metrics, BLEU\footnote{\url{https://github.com/moses-smt/mosesdecoder/}} and ROUGE\footnote{\url{https://pypi.org/project/rouge}}, as well as normalized tree edit distance (NTED) to measure the overall quality of the generated instruction texts with respect to the ground-truth instructions. For NTED, we employ similar procedures used in Chang et al. \cite{chang2018recipescape}, i.e., we represent instructions as a tree structure where the verbs and nouns (extracted by spaCy) are respectively on stems and leaves representing the relation of actions (i.e. cooking methods) and subjects (i.e. ingredients and cooking tools). We then utilize Zhang-Shasha algorithm to score the edit distance which counts the numbers of INSERT, REMOVE, and REPLACE operations required to get the gold reference from the generated instruction. Lastly, we normalize the tree edit distance by the total number of nodes in both generated and reference trees.

\textbf{Between-field ingredient coherence}
Typically, we expect that all ingredient words should be consistently referred to across all fields. To investigate how well the model captures this phenomenon, we measure the overlap between a set of ingredient words extracted from the generated instructions ($I_g$) and those in the input ingredient texts ($I$) using Jaccard similarity. First, we extract the ingredient-containing root nouns from the generated instructions and the input ingredient texts. We use two extraction methods for those root nouns as illustrated in Figure \ref{fig:spacy}. The first method has been applied previously in the F1 calculation. The second method extracts root nouns from instructions using an ingredient dictionary to filter out non-ingredients. The dictionary is derived from~\cite{weber2016insights} with 89 additional ingredients added by a member of the research team to increase ingredients coverage with respect to Recipe1M dataset. In total, the vocabulary contains 1,992 root nouns. Then, we compute Jaccard similarity scores between the two sets of root nouns. To establish a baseline, we also perform the same procedures to compare the ingredient coherence between human-written instructions ($I_h$) and human-written ingredients ($I$).

\subsection{Results} \label{sec:compare}
\textbf{Ingredient and instruction generations.} 
Overall, the results confirm the effectiveness of RecipeGPT in generating ingredient and cooking instruction texts. Table \ref{tbl:performance} displays the model performances on different sampling hyperparameters ($k$) for top-k sampling. The best model is subsequently selected for live deployment.
In Table~\ref{tbl:performance}(a), we experiment with different $k$ to examine its effects on the generation diversity and quality. As we can see, models with lower $k$ consistently perform better across all evaluation metrics. Next, $k$ is also positively associated with the length of generation as indicated by the average number of ingredients and the brevity penalty. Ultimately, we set $k$ to 3 in the live deployment to achieve a good balance of quality and diversity. 

We also evaluate different training approaches as presented in Table~\ref{tbl:performance}(b).
As we can see, all models show similar performance levels, suggesting that the knowledge captured in the pre-trained weights are not that helpful when training with a large dataset. Next, fine-tuning on a more complex model (355M vs. 124M) produces a superior model according to perplexity, which is consistent with previous findings \cite{zellers2019defending,zhang2019dialogpt}.
As the generation results among those models are indistinguishable and it takes approximately 9 seconds for 124M and 12 seconds for 355M to process a test case, we deploy the 117M fine-tuned GPT-2 in the live version of RecipeGPT\footnote{Using a learning rate of 1e-4, the deployed model in our system takes 5 days, approximately 5 epochs, to converge in a single NVIDIA Tesla V100 GPU.}.

\textbf{Ingredient coherence.} 
The Jaccard similarity scores between ($I_g$, $I$) and ($I_h$, $I$) are 0.53 and 0.49, respectively. This suggests that RecipeGPT captures the between-field ingredient coherence as well as human authors, if not better. Surprisingly, the score for human-written instructions is slightly lower than that of the generated instructions. Upon further inspection, we found that human authors sometimes refer to the input ingredients as a whole when writing the instructions instead of mentioning each individual ingredients, e.g., "combine all ingredients" v.s. "combine vodka and orange juice", thus lowering the overlap.

\begin{table}[!tp]
\caption{Model Performances}
\label{tbl:performance}
\scalebox{0.75}{
\begin{threeparttable}
\begin{tabular}{l|ll|llll|l}
\toprule
Generation output &\multicolumn{2}{c|}{Ingredients} &  \multicolumn{4}{c|}{Instructions} & \\
\midrule
  & F1 & \# Ingr. & BLEU & BP & R-L & NTED & PPL  \\
\midrule
\multicolumn{7}{l}{(a) Performances of fine-tuned GPT-2 (124M) on validation set} \\
\midrule
Top-k sampling with k = 1 & 0.79 & 7.6  & 9.81 & 0.62 & 0.39 & 0.51 & \\
\hspace{8.5em} k = 3 & 0.76 & 7.9 & 8.29 & 0.70 & 0.37 & 0.52 & \\
\hspace{8.5em} k = 5  & 0.7 & 8.0 & 7.81 & 0.75 & 0.36 & 0.53 & \\
\hspace{8.5em} k = 10 & 0.74 & 8.3 & 7.42 & 0.83 & 0.35 & 0.53 & \\
\hspace{8.5em} k = 30  & 0.71 & 8.7 & 7.15 & 0.94 & 0.34 & 0.54 & \\
\midrule
\multicolumn{7}{l}{(b) Performances of different RecipeGPT models on test set} \\
\midrule
Trained from scratch (124M)& 0.75 & 7.6  & 8.58 & 0.71 & 0.37 & 0.52 & 3.77 \\
Fine-tuned GPT-2 (124M)  & 0.76 & 7.8 & 8.34 & 0.71 & 0.36 & 0.52 & 3.70 \\ 
Fine-tuned GPT-2 (355M) & 0.77 & 7.9 & 8.29 & 0.70 & 0.37 & 0.52 & 3.63\\
\bottomrule
\end{tabular}
\begin{tablenotes}
\item F1: F1 of two sets of lemmatized root nouns, \# Ingr: average number of ingredients, BP: Brevity Penalty estimated in BLEU, R-L: ROUGE-L, NTED: Normalized Tree Edit Distance, PPL: BPE Perplexity
\end{tablenotes}
\end{threeparttable}}
\end{table}

\section{Demonstration \& Conclusion} \label{sec:conclusion}

\textbf{Demonstration.} During the demo session, we will demonstrate the whole recipe generation and evaluation process as shown in Figure~\ref{fig:userflow}. Both cooking instruction generation and ingredient generation tasks will be available for demonstration. For each task, users will provide specific recipe contexts (e.g., recipe title and ingredients/cooking instructions) to generate respective recipe texts. Single recipe evaluation features are also available to use. RecipeGPT is accessible online at ~\url{https://recipegpt.org/} 
\footnote{We share the code used to run the experiments at \url{https://github.com/LARC-CMU-SMU/RecipeGPT-exp}.}

\textbf{Conclusion.} RecipeGPT is a novel generative pre-trained transformer based recipe generation and evaluation system. The evaluations illustrate its potential in automatic recipe generation. As the system is publicly accessible online, we hope that this will encourage others to try experimenting with different recipe contexts to further investigate its potentials and behaviors.
Lastly, we also provide several user-interface features in RecipeGPT to assist users in examining the quality of the generation at the recipe level, and suggest potential ways to improve recipe generation models.


\section{Acknowledgement}
This research is supported by the National Research Foundation, Prime Minister's Office, Singapore under its International Research Centres in Singapore Funding Initiative.

\bibliographystyle{ACM-Reference-Format}
\bibliography{main}


\begin{thebibliography}{10}


\ifx \showCODEN    \undefined \def \showCODEN     #1{\unskip}     \fi
\ifx \showDOI      \undefined \def \showDOI       #1{#1}\fi
\ifx \showISBNx    \undefined \def \showISBNx     #1{\unskip}     \fi
\ifx \showISBNxiii \undefined \def \showISBNxiii  #1{\unskip}     \fi
\ifx \showISSN     \undefined \def \showISSN      #1{\unskip}     \fi
\ifx \showLCCN     \undefined \def \showLCCN      #1{\unskip}     \fi
\ifx \shownote     \undefined \def \shownote      #1{#1}          \fi
\ifx \showarticletitle \undefined \def \showarticletitle #1{#1}   \fi
\ifx \showURL      \undefined \def \showURL       {\relax}        \fi
\providecommand\bibfield[2]{#2}
\providecommand\bibinfo[2]{#2}
\providecommand\natexlab[1]{#1}
\providecommand\showeprint[2][]{arXiv:#2}

\bibitem[\protect\citeauthoryear{Bosselut, Levy, Holtzman, Ennis, Fox, and
  Choi}{Bosselut et~al\mbox{.}}{2017}]%
        {bosselut2017simulating}
\bibfield{author}{\bibinfo{person}{Antoine Bosselut}, \bibinfo{person}{Omer
  Levy}, \bibinfo{person}{Ari Holtzman}, \bibinfo{person}{Corin Ennis},
  \bibinfo{person}{Dieter Fox}, {and} \bibinfo{person}{Yejin Choi}.}
  \bibinfo{year}{2017}\natexlab{}.
\newblock \showarticletitle{Simulating action dynamics with neural process
  networks}.
\newblock \bibinfo{journal}{\emph{arXiv preprint arXiv:1711.05313}}
  (\bibinfo{year}{2017}).
\newblock


\bibitem[\protect\citeauthoryear{Chang, Guillain, Jung, Hare, Kim, and
  Agrawala}{Chang et~al\mbox{.}}{2018}]%
        {chang2018recipescape}
\bibfield{author}{\bibinfo{person}{Minsuk Chang},
  \bibinfo{person}{L{\'e}onore~V Guillain}, \bibinfo{person}{Hyeungshik Jung},
  \bibinfo{person}{Vivian~M Hare}, \bibinfo{person}{Juho Kim}, {and}
  \bibinfo{person}{Maneesh Agrawala}.} \bibinfo{year}{2018}\natexlab{}.
\newblock \showarticletitle{Recipescape: An interactive tool for analyzing
  cooking instructions at scale}. In \bibinfo{booktitle}{\emph{Proceedings of
  the 2018 CHI Conference on Human Factors in Computing Systems}}. ACM,
  \bibinfo{pages}{451}.
\newblock


\bibitem[\protect\citeauthoryear{Majumder, Li, Ni, and McAuley}{Majumder
  et~al\mbox{.}}{2019}]%
        {majumder2019generating}
\bibfield{author}{\bibinfo{person}{Bodhisattwa~Prasad Majumder},
  \bibinfo{person}{Shuyang Li}, \bibinfo{person}{Jianmo Ni}, {and}
  \bibinfo{person}{Julian McAuley}.} \bibinfo{year}{2019}\natexlab{}.
\newblock \showarticletitle{Generating Personalized Recipes from Historical
  User Preferences}. In \bibinfo{booktitle}{\emph{EMNLP}}.
  \bibinfo{pages}{5975--5981}.
\newblock
\urldef\tempurl%
\url{https://www.aclweb.org/anthology/D19-1613}
\showURL{%
\tempurl}


\bibitem[\protect\citeauthoryear{Marin, Biswas, Ofli, Hynes, Salvador, Aytar,
  Weber, and Torralba}{Marin et~al\mbox{.}}{2019}]%
        {marin2019recipe1m}
\bibfield{author}{\bibinfo{person}{Javier Marin}, \bibinfo{person}{Aritro
  Biswas}, \bibinfo{person}{Ferda Ofli}, \bibinfo{person}{Nicholas Hynes},
  \bibinfo{person}{Amaia Salvador}, \bibinfo{person}{Yusuf Aytar},
  \bibinfo{person}{Ingmar Weber}, {and} \bibinfo{person}{Antonio Torralba}.}
  \bibinfo{year}{2019}\natexlab{}.
\newblock \showarticletitle{Recipe1m+: A dataset for learning cross-modal
  embeddings for cooking recipes and food images}.
\newblock \bibinfo{journal}{\emph{IEEE transactions on pattern analysis and
  machine intelligence}} (\bibinfo{year}{2019}).
\newblock


\bibitem[\protect\citeauthoryear{Radford, Wu, Child, Luan, Amodei, and
  Sutskever}{Radford et~al\mbox{.}}{2019}]%
        {radford2019language}
\bibfield{author}{\bibinfo{person}{Alec Radford}, \bibinfo{person}{Jeffrey Wu},
  \bibinfo{person}{Rewon Child}, \bibinfo{person}{David Luan},
  \bibinfo{person}{Dario Amodei}, {and} \bibinfo{person}{Ilya Sutskever}.}
  \bibinfo{year}{2019}\natexlab{}.
\newblock \showarticletitle{Language models are unsupervised multitask
  learners}.
\newblock \bibinfo{journal}{\emph{OpenAI Blog}} \bibinfo{volume}{1},
  \bibinfo{number}{8} (\bibinfo{year}{2019}).
\newblock


\bibitem[\protect\citeauthoryear{Salvador, Drozdzal, Giro-i Nieto, and
  Romero}{Salvador et~al\mbox{.}}{2019}]%
        {salvador2019inverse}
\bibfield{author}{\bibinfo{person}{Amaia Salvador}, \bibinfo{person}{Michal
  Drozdzal}, \bibinfo{person}{Xavier Giro-i Nieto}, {and}
  \bibinfo{person}{Adriana Romero}.} \bibinfo{year}{2019}\natexlab{}.
\newblock \showarticletitle{Inverse cooking: Recipe generation from food
  images}. In \bibinfo{booktitle}{\emph{Proceedings of the IEEE Conference on
  Computer Vision and Pattern Recognition}}. \bibinfo{pages}{10453--10462}.
\newblock


\bibitem[\protect\citeauthoryear{Varshney, Pinel, Varshney, Bhattacharjya,
  Sch{\"o}rgendorfer, and Chee}{Varshney et~al\mbox{.}}{2019}]%
        {varshney2019big}
\bibfield{author}{\bibinfo{person}{Lav~R Varshney}, \bibinfo{person}{F Pinel},
  \bibinfo{person}{KR Varshney}, \bibinfo{person}{D Bhattacharjya},
  \bibinfo{person}{A Sch{\"o}rgendorfer}, {and} \bibinfo{person}{Y-M Chee}.}
  \bibinfo{year}{2019}\natexlab{}.
\newblock \showarticletitle{A big data approach to computational creativity:
  The curious case of Chef Watson}.
\newblock \bibinfo{journal}{\emph{IBM Journal of Research and Development}}
  \bibinfo{volume}{63}, \bibinfo{number}{1} (\bibinfo{year}{2019}),
  \bibinfo{pages}{7--1}.
\newblock


\bibitem[\protect\citeauthoryear{Weber and Achananuparp}{Weber and
  Achananuparp}{2016}]%
        {weber2016insights}
\bibfield{author}{\bibinfo{person}{Ingmar Weber} {and}
  \bibinfo{person}{Palakorn Achananuparp}.} \bibinfo{year}{2016}\natexlab{}.
\newblock \showarticletitle{Insights from machine-learned diet success
  prediction}. In \bibinfo{booktitle}{\emph{Biocomputing 2016: Proceedings of
  the Pacific Symposium}}. World Scientific, \bibinfo{pages}{540--551}.
\newblock


\bibitem[\protect\citeauthoryear{Zellers, Holtzman, Rashkin, Bisk, Farhadi,
  Roesner, and Choi}{Zellers et~al\mbox{.}}{2019}]%
        {zellers2019defending}
\bibfield{author}{\bibinfo{person}{Rowan Zellers}, \bibinfo{person}{Ari
  Holtzman}, \bibinfo{person}{Hannah Rashkin}, \bibinfo{person}{Yonatan Bisk},
  \bibinfo{person}{Ali Farhadi}, \bibinfo{person}{Franziska Roesner}, {and}
  \bibinfo{person}{Yejin Choi}.} \bibinfo{year}{2019}\natexlab{}.
\newblock \showarticletitle{Defending Against Neural Fake News}.
\newblock \bibinfo{journal}{\emph{arXiv preprint arXiv:1905.12616}}
  (\bibinfo{year}{2019}).
\newblock


\bibitem[\protect\citeauthoryear{Zhang, Sun, Galley, Chen, Brockett, Gao, Gao,
  Liu, and Dolan}{Zhang et~al\mbox{.}}{2019}]%
        {zhang2019dialogpt}
\bibfield{author}{\bibinfo{person}{Yizhe Zhang}, \bibinfo{person}{Siqi Sun},
  \bibinfo{person}{Michel Galley}, \bibinfo{person}{Yen-Chun Chen},
  \bibinfo{person}{Chris Brockett}, \bibinfo{person}{Xiang Gao},
  \bibinfo{person}{Jianfeng Gao}, \bibinfo{person}{Jingjing Liu}, {and}
  \bibinfo{person}{Bill Dolan}.} \bibinfo{year}{2019}\natexlab{}.
\newblock \showarticletitle{DIALOGPT: Large-Scale Generative Pre-training for
  Conversational Response Generation}.
\newblock \bibinfo{journal}{\emph{arXiv preprint arXiv:1911.00536}}
  (\bibinfo{year}{2019}).
\newblock


\end{thebibliography}

\end{document}